\documentclass[letterpaper, 10 pt, conference]{ieeeconf}  
\pdfoutput=1
\IEEEoverridecommandlockouts                              
\usepackage{graphics} 
\usepackage{epsfig} 
\usepackage{times} 
\usepackage{amsmath} 
\usepackage{amssymb}  
\usepackage{algorithm}
\usepackage[noend]{algpseudocode}
\usepackage{xcolor}
\usepackage{balance}
\usepackage{caption}
\usepackage{subcaption}

\DeclareMathOperator*{\argmax}{\arg\!\max}

\title{\LARGE \bf
Meta-Learning for Multi-objective Reinforcement Learning
}

\author{Xi Chen$^{1}$, Ali Ghadirzadeh$^{1,2}$, M{\aa}rten Bj{\"o}rkman$^{1}$ and Patric Jensfelt$^{1}$
\thanks{
\newline
$^{1}$RPL, KTH Royal Institute of Technology, Sweden \newline
$^{2}$Intelligent Robotics research group, Aalto University, Finland \newline
{\tt\small [xi8][algh][celle][patric]@kth.se}}
}

\begin{document}

\maketitle
\thispagestyle{empty}
\pagestyle{empty}

\begin{abstract}
Multi-objective reinforcement learning (MORL) is the generalization of standard reinforcement learning (RL) approaches 
to solve sequential decision making problems that consist of several, possibly conflicting, objectives. 
Generally, in such formulations, there is no single optimal policy which optimizes all the objectives simultaneously, and instead, a number of policies has to be found each optimizing a preference of the objectives.  
In this paper, we introduce a novel MORL approach by training a meta-policy, a policy simultaneously trained with multiple tasks sampled from a task distribution, for a number of randomly sampled Markov decision processes (MDPs). 
In other words, the MORL is framed as a meta-learning problem, with the task distribution given by a distribution over the preferences. 
We demonstrate that such a formulation results in a better approximation of the Pareto optimal solutions in terms of both the optimality and the computational efficiency. 
We evaluated our method on obtaining Pareto optimal policies using a number of continuous control problems with high degrees of freedom.
\end{abstract}

\section{INTRODUCTION}

Reinforcement learning (RL) is a framework to train an agent to acquire a behavior by reinforcing actions that maximize a notion of task-relevant future rewards.
A reward function, i.e., the function that assigns a reward value to every action-decision made by the agent, 
is designed to guide the training to implement the behavior. 
For specific applications, there are learning frameworks that train with sparse reward
\cite{riedmiller2018learning,andrychowicz2017hindsight, ghadirzadeh2017deep}. However, in many real world problems, a behavior is best realized by finding a trade-off among a number of conflicting reward functions.
This is known as multi-objective reinforcement learning (MORL), which is a fundamental problem in designing many autonomous systems.
As an example, real systems consume energy and, in most cases, reducing energy consumption as a desirable behavior lowers the performance of the system. 
Other examples are safety and stability which   contradict the main objectives of the system in many cases. 

In many RL problems, the trade-off between different objectives is found by constructing a synthetic reward function, e.g., a weighted linear sum of the objectives, using expert knowledge \cite{schulman2017proximal,duan2016benchmarking}.
This approach may not be suitable for problems that, 
(1) aim at realizing a multifaceted behavior consisting of several phases in sequence, therefore, preventing utility of a set of predetermined preferences over the objectives, 
(2) are required to adapt itself to operate in different modes, e.g., to operate in energy-saving versus high-performance mode, and
(3) for which it is hard to find a set of suitable preference over objectives manually, a problem known as preference elicitation (PE) \cite{liu2015multiobjective}.

\begin{figure}[thpb]
  \centering
  \includegraphics[scale=0.27]{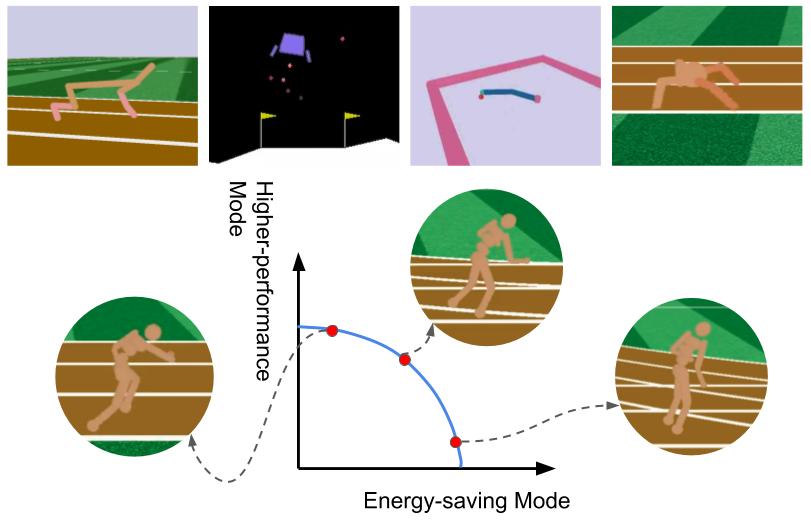}
  \caption{The five continues control tasks that are used in our experiments.
  }
  \label{fig:overview}
\end{figure}

However, most of the current MORL solutions are limited to simple toy examples with discrete state-action spaces \cite{liu2015multiobjective, vamplew2011empirical, roijers2013survey}. 
In this work, we introduce a novel approach to train deep policies for complex continuous control tasks based on meta-learning \cite{finn2017model}.
We train a meta-policy with tasks sampled from a distribution over MDPs with different preferences over the objectives. 
We demonstrate that our proposed approach outperforms direct policy search methods, e.g., \cite{parisi2014policy}, at obtaining Pareto optimal policies for different continuous RL problems. 
Given a reward vector, a policy is Pareto optimal w.r.t. the expected return measure if it results in higher expected return for at least one component of the vector compared to all other policies.

The rest of this paper is organized as following: 
We outline related work in Section \ref{sec:relatedwork}. 
We introduce required background in Section \ref{sec:preliminaries}.
Section \ref{sec:meta-morl} provides the details of our method. In Section \ref{sec:experiments}, we explain our experimental results. Finally, in Section \ref{sec:conclusion}, we conclude the paper and introduce our future work.

\section{RELATED WORK}
\label{sec:relatedwork}
MORL problems can be viewed as the combination of multi-objective optimization (MOO) and RL techniques to find Pareto optimal solutions which are defined as non-dominated solutions representing trade-offs among multiple conflicting objectives \cite{liu2015multiobjective, vamplew2011empirical, roijers2013survey}.
MORL methods can be divided into two main categories: single- and multi-policy approaches, with the main difference being the number of policies that are learned at each run.
Single-policy algorithms 
transform the multi-objective into a single-objective problem using a parametric scalarization function \cite{miettinen2002scalarizing}, \cite{miettinen2001some}. 
The solution is then found by  running
the  RL  method  several  times  with  different  parameters.
In  contrast,  multi-policy  algorithms  update  several  policies simultaneously  in  each  run  to  improve  the  current  Pareto optimal solution. 
Evolutionary methods, e.g., \cite{coello2007evolutionary, konak2006multi, zitzler2001spea2} and \cite{deb2000fast} are good examples of the latter category.

In case a scalar value can be computed from the multiple objectives using domain-specific expert knowledge, the problem can be converted to a single objective RL problem which can be addressed with standard RL solutions. As examples, Schulman et al., \cite{schulman2017proximal} and Duan et al., \cite{duan2016benchmarking} applied policy search methods to a number of continuous control benchmarks by hand-designing a synthetic reward function from multiple objectives 
for different locomotion tasks.  
However, setting the parameters a priori requires fine-tuning and it is non-intuitive to find the right trade-off \cite{liu2015multiobjective,brys2017multi}.
Lizotte et al., \cite{lizotte2010efficient} proposed a multi-objective Q-learning solution by linearly combining the objectives related to symptoms and side-effects of different treatment plans. They identified dominated actions, as well as, preferred actions at different areas of the objective space by using a tabular Q-function with an extra entry for the objectives. 
However, it is well-known that multi-objective Q-learning by the weighted-sum approach cannot approximate concave parts of the true Pareto front \cite{das1997closer}. 
Van Moffaert et al., \cite{van2013scalarized} introduced a similar multi-objective Q-learning method but based on the Chebyshev scalarization function as a replacement for the weighted-sum to approximate the Pareto front.
Still the method updates the Q-function based on action-decisions made according to a preference over the objectives which is encoded by the weights of the Chebyshev function, resulting in sub-optimal Pareto fronts. 
In a later study \cite{van2013hypervolume}, they improved the method by exploiting a hypervolume-based action-selection mechanism. The hypervolume is a measure used for performance assessment of the Pareto optimal solutions and not a single preference given by a set of weights.  
However, estimation of the hypervolume is a computationally expensive NP-hard problem \cite{van2013hypervolume}.
Our approach trains stochastic policies with a parametric scalarization method by sampling different random weights during the meta-learning phase to alleviate the aforementioned issue regarding the sub-optimality of the Pareto front, while still performing tractable computations. 
The work done by Natarajan and Tadepalli \cite{natarajan2005dynamic} resembles our work, in that it also tries to obtain a number of policies,  based on which a non-dominated Pareto front can be established for preferences over different objectives encoded by a weight vector. 
In their work, a bag of policies is constructed recursively to approximate the front, as well as to initialize new policies such that, the policy training can be resumed from the closest policy in the bag.
We instead propose to initialize new policies, in a so-called adaptation phase, from the updated meta-policy which is trained with many different weight values.  
The meta-policy is proven to adapt to new conditions faster \cite{finn2017model}, and can be realized by both stochastic and deterministic policies. 

Recently, policy search approaches have been studied in the context of MORL. 
Parisi et al., \cite{parisi2014policy} proposed to estimate a number of Pareto optimal policies by performing gradient ascent in the policy parameter space by solving different single-objective RL problems found by different convex combinations of the objectives. 
In contrast, our method does not initialize each policy randomly and it obtains a suitable initial policy parameter in the meta-learning phase, 
resulting in a more efficient way of estimating the Pareto optimal policies. 
Furthermore, as we demonstrate in the experiment section, our method outperforms \cite{parisi2014policy} for complicated tasks requiring a hierarchy of skills to be obtained in sequences. 

In case of multi-policy approaches, 
H. Handa \cite{handa2009eda} proposed the EDA-RL method which applies the Estimation of Distribution (EDA) evolutionary algorithms to solve RL problems by searching in the policy parameter space. He further extended EDA-RL
to MORL with a Pareto dominance based fitness metric \cite{handa2009solving}. 
Pirotta et al., \cite{pirotta2015multi} introduced a manifold-based policy search MORL solution which assumes policies to be sampled from a manifold, i.e., a parametric distribution over the policy parameter space. They proposed to update the manifold according to an indicator function, such that the sampled policies construct an improved Pareto front. 
In a more recent work, Parisi et al., \cite{parisi2017manifold}  extended the method to work with the hypervolume indicator function as a more well-suited measure when evaluating optimality of different Pareto fronts. 
However, the main issue of these approaches is that 
the number of parameters grow quadratically with the number of policy parameters. As an example, a policy with 30 parameters requires 7448 parameters to model the manifold \cite{parisi2017manifold} with Gaussian mixture models. 
As the result, these approaches may not be applicable to train deep policies with several thousands of parameters. 

In this paper, we propose to frame policy search MORL for continuous control with large numbers of policy parameters, e.g., deep policies, as a meta-learning problem. 
In contrast to earlier approaches
We train a meta-policy to estimate the Pareto front implicitly using different tasks sampled from a distribution of the preferences over the objectives, such that, we can obtain the Pareto optimal solution of a given preference by fine-tuning the meta-policy with a few gradient updates, the Pareto front can be constructed more efficiently


\section{PRELIMINARIES}
\label{sec:preliminaries}

A multi-objective sequential decision making process can be represented by a Markov decision process (MDP) which is defined by the tuple $(\mathcal{S}, \mathcal{A}, \mathcal{P}, r, p(s_0), \gamma)$, 
where $\mathcal{S}$ is the state space (continuous), 
$\mathcal{A}$ is the action space (continuous), 
$\mathcal{P}:\mathcal{S}\times\mathcal{A}\times\mathcal{S} \rightarrow [0, \infty )$ is the action-conditioned state transition distribution, 
$r:\mathcal{S} \times \mathcal{A} \rightarrow \mathbb{R}^q$ represents the reward functions corresponding to $q$ different objectives, 
$p(s_0):\mathcal{S}\rightarrow[0, \infty )$ is the distribution of initial states, and $\gamma\in[0,1]$ is a discount factor.
We consider episodic tasks in which an episode starts by sampling an initial state from $p(s_0)$, and then for every time-step $t < H$, sampling actions $a_t$ from $\pi_\theta(a_t|s_t)$, a parametric stochastic policy. The successor state in each time step is given according to $p(s_{t+1}|s_t, a_t)$ and a reward vector $r_t = r(s_t, a_t)$ is provided at every time-step by the environment.
For every state-action pair in the episode, the return is a vector defined as the discounted sum of future rewards, $R_t = \sum_{t'=t} \gamma^{t'-t} r(s_{t'}, a_{t'})$.

\section{Multi-objective policy search }
\label{sec:mops}
Reinforcement learning is based on performing exploratory actions and reinforcing the actions that result in return outcomes exceeding the expectation of the agent. 
The expectation of the agent is modeled by a state value function, represented by a neural network in our work, and is continually updated to model the expected return for a given state $s$ while following the policy $\pi_\theta$,
\begin{equation*}
V^\pi(s) = \mathbb{E}[\sum_{t=0}^T \gamma^t r_t | s_0 = s]. 
\end{equation*}

The agent performs exploratory actions and compare the actual return outcomes for every state-action pair with the expected value to form the advantage function,
\begin{equation*}
A(s_t, a_t) = R_t - V^\pi(s_t).
\end{equation*}
Here, the advantage is a vector which can be converted to a scalar value by a parametric scalaraization function $f_\omega(A)$, e.g., the Chebyshev function $f_\omega(A)=(\sum_i^q \omega_i |A_i - z_i|^p)^\frac{1}{p}$ ($1\le p < \infty$ and $z$ is a utopian point) for concave problems, or weighted sum, $f_\omega(A) =\sum_{i=1}^q \omega_i A_i$, for convex optimzations.  
The policy parameters are updated such that state-action pairs with $f_\omega(A) > 0$ are reinforced, i.e., become more probable in the future. 
This can be achieved by minimizing a loss function $\mathcal{L}(\theta, \omega)$, 
in our case the clipped version of TRPO loss \cite{schulman2017proximal}, known as PPO  \cite{schulman2015trust},
\begin{equation*}
\mathcal{L}(\theta,\omega) = \mathbb{E}_\tau[\frac{ \pi_{\theta}(a_t|s_t)}{\pi_{\theta_{old}}(a_t|s_t)}f_\omega(A_t)-\beta D_{KL}(\pi_{\theta_{old}} || \pi_\theta)],
\label{eq:trpo_loss}
\end{equation*}
where, $D_{KL}$ represents Kullback Leibler (KL)-divergence and $\tau$ represents the state-action trajectories and $\beta$ is a scalar parameter found empirically. The parameters are updated such that the new policy assigns higher probability to state-action pairs resulting in positive scalarized advantages while the assigned policy distribution is not drastically changed from the old policy $\pi_{\theta_{old}}$.

\section{Multi-objective Reinforcement Learning}
\label{sec:meta-morl}

Our approach to solve MORL problems is to train a policy which can be quickly adapted to optimize an unseen objective function constructed by a new preference vector (arbitrary non-negative $\omega$s such that $\sum_i \omega_i = 1$). The policy is called a \textit{meta-policy} and is trained such that a few more RL iterations would be sufficient to optimize the given objective function. 
The meta-policy, once it is trained, does not  contribute to the construction of the Pareto front by itself. But it is used as an optimal initial policy to efficiently train for the different objectives to construct the Pareto front.  

During the training, the meta-policy $\pi_\theta$ is updated by multiple objectives constructed by different arbitrary preference vectors. The rule of the training is to ensure few-shot learning for every objective used for the policy training. In other words, every training objective should be maximized after a few number of  RL iterations. At the test time, once an unseen objective is provided, the meta-policy can be efficiently optimized in case the objective is sufficiently close to the objectives used during the training. 

A popular formulation for few-shot learning, is the model-agnostic meta-learning (MAML) method \cite{finn2017model}, in which the meta-policy parameters are explicitly updated to be a few gradients away from the optimal parameters for every objective function.
Without loss of generality, MAML can be suitably integrated to our MORL framework, though it is not the only option. In principle, other few-shot learning methods may also be used to train the meta-policy and to obtain the Pareto optimal solutions. 

Our method consists of three steps, 
(1) an adaptation phase, in which a number of policies are updated for a few number of iterations from the meta-policy, (2) a meta-policy training phase, in which the meta-policy is updated by aggregating data generated by the policies trained in the previous phase, and (3) a fine-tuning phase, in which the Pareto optimal policies are trained after being initialized by the meta-policy parameters.
In the following sections, different phases of the training are introduced in more details.

\subsection{Adaptation phase}
In the adaptation phase, a number of preference vectors $\omega_i$ are randomly sampled ($\omega_i \ge 0, \forall i$ and $\sum_i \omega_i  = 1.0$). For each preference vector, a policy is initialized with the parameters of the meta-policy.
The policies are updated for one iteration using state-action trajectories generated by running the meta-policy, and  return values found according to the assigned task (specified by $\omega_i$).  
In short, each policy $\pi_{\theta_i}$ is trained as, 
\begin{equation}
\theta_i = \argmax_{\theta'}{ \mathcal{L}(\theta',\omega_i)}|_{\theta_{init} = \theta},
\label{eq:one_step_update}
\end{equation}
where, $\theta$ is the meta-policy parameter. 

\subsection{Meta-policy training phase}

The meta-policy is updated in the meta-learning phase by aggregating the information of the policies $\pi_{\theta_i}$ trained in the previous step, 

\begin{equation}
\theta = \argmax_{\theta'} {\sum_i \mathcal{L}(\theta',\omega_i)}|_{\theta_{init} = \theta},
\label{eq:meta_learning}
\end{equation}
that is the first-order approximation of the MAML.  
\subsection{Fine-tuning phase}
Finally, once the meta-policy is trained, a set of Pareto optimal policies is found by fine-tuning the meta-policy for a number of iterations with different preference vectors. 

\begin{algorithm}
  \caption{Meta-MORL policy training} 
  \textbf{Input} $p(\mathbf{\omega})$: the distributions over the preference vector 
  \label{alg:meta_training}
  \begin{algorithmic}[1]
    \State Randomly initialize the meta-policy $\pi_\theta$ 
    \While{training $\pi_\theta$}
        \State Sample trajectories $\mathcal{D}$ following $\pi_\theta$ 
      	\For{each task $i$}
      	    \State Sample $\omega_i \sim  p(\omega)$
      		\State Update $\pi_{\theta_i}$ with $\mathcal{D}$ according to Eq.~\ref{eq:one_step_update}
      		\State Sample trajectories $\mathcal{D}_i$ following $\pi_{\theta_i}$
      	\EndFor
      	\State Update meta-policy with $\mathcal{D}_i$, $\omega_i$ according to Eq.~\ref{eq:meta_learning}
    \EndWhile
    \State Construct the Pareto front by the fine-tuning phase
  \end{algorithmic}
\end{algorithm}

The details of our method are provided in Algorithm 1. The meta-policy is initialized randomly and it is updated in the training loop. 
In each iteration, trajectories of states, actions and rewards are sampled by running the meta-policy. 
For every task, a weight vector is sampled and a new policy is trained, in the adaptation phase. 
The trained policies are used to sample new trajectories based on which the meta-policy is updated in the meta-learning phase.
Finally, a set of Pareto front policies is obtained by initializing the policies with the meta-policy and training them for a number of iterations. 

Our method resembles the multi-policy approaches, in that it optimizes a meta-policy to estimate the Pareto front implicitly,
but it also resembles the single-policy approaches, since the final Pareto front policies are trained individually for each objective during the fine-tuning phase. 
Our method is not limited by the number of parameters in the policy and it is more efficient to find the optimal policy for each objective function.

\section{EXPERIMENT}
\label{sec:experiments}
\begin{figure*}
\begin{subfigure}{0.32\textwidth}
\centering
  \includegraphics[scale=0.42]{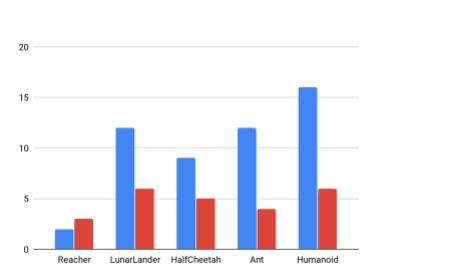}
\caption{The number of non-dominated policies found by the two methods} \label{fig:non_dominated_p}
\end{subfigure}
\hspace*{\fill} 
\begin{subfigure}{0.32\textwidth}
\centering
  \includegraphics[scale=0.42]{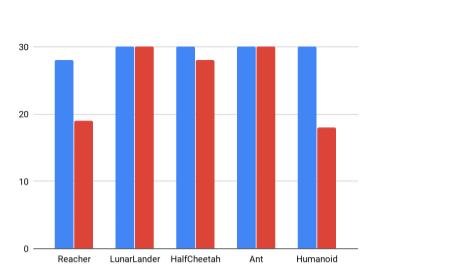}
\caption{The number of valid policies found by the two methods} \label{fig:valid_p}
\end{subfigure}
\hspace*{\fill} 
\begin{subfigure}{0.32\textwidth}
\centering
  \includegraphics[scale=0.42]{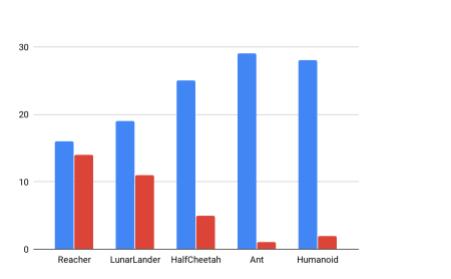}
\caption{The number of policies that achieves higher scalarized return given a weight vector.} \label{fig:higher_r}
\end{subfigure}
\caption{Different policies found by the two methods. The blue bars denote the results of our method and the red bars denote the result of RA.} \label{fig:1}
\end{figure*}


\begin{figure*}[thpb]
  \centering
  \includegraphics[scale=0.36]{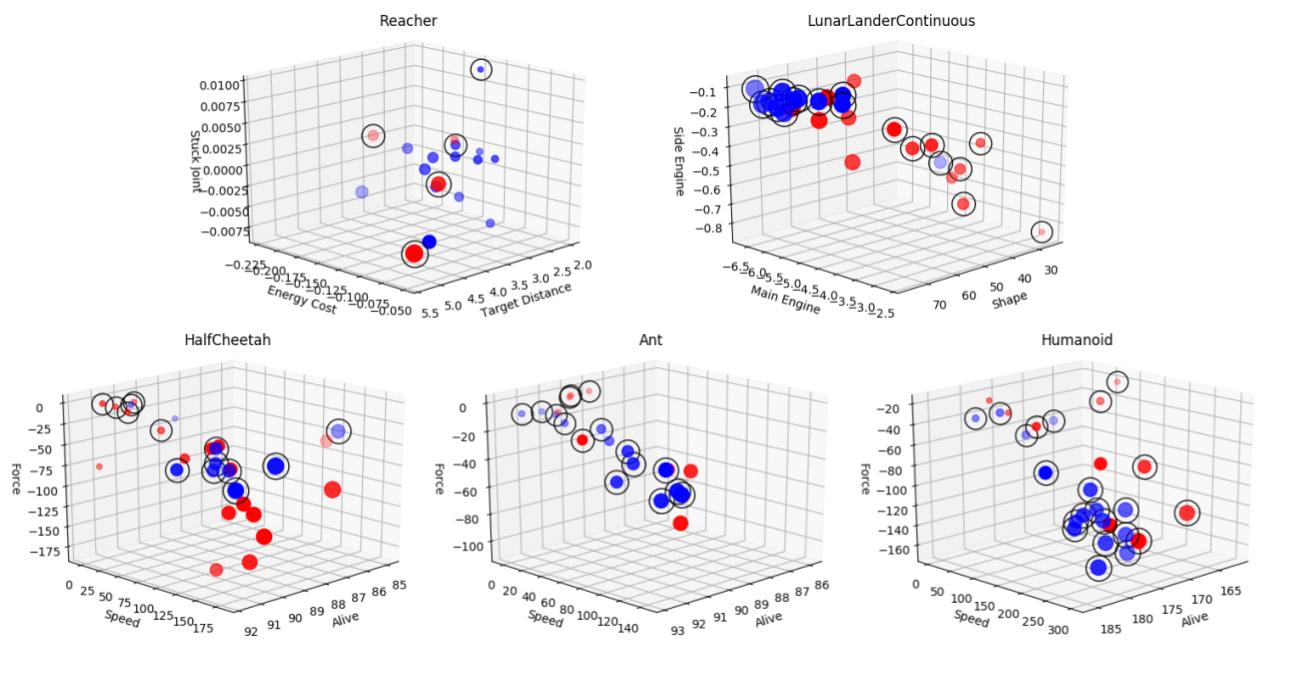}
    \caption{The Pareto optimal policies found by the two methods. For tasks with more than three objectives, we only illustrate the three most informative ones. The blue points denote the policies found by our method, and the red points denote the policies found by the baseline. The points with black circle denote the non-dominated policies. The size of points is proportional to the distance to the origin to better visualize in $3D$.}
    \label{fig:pf_plots}
\end{figure*}

\begin{figure*}[thpb]
  \centering
  \includegraphics[scale=0.38]{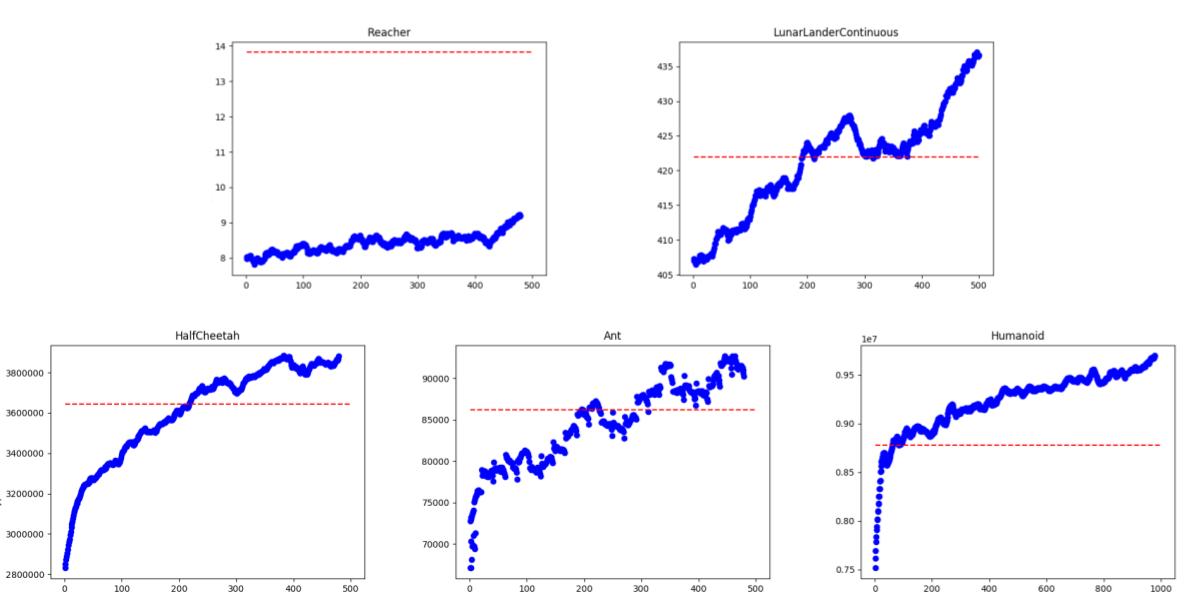}
    \caption{The improvements of the hypervolume indicator (vertical axis) with respect to the iteration of fine-tuning the meta policy (horizontal axis). The blue curve denotes the hypervolume and the red line denotes the final hypervolume of the Pareto front estimated by RA. }
    \label{fig:hv_indicator}
\end{figure*}

In the experiments, we aim to answer the following questions:
(1) does the proposed MORL solution successfully train a set of policies with a large number of parameters that estimate the Pareto front for continuous control problems? 
(2) compared to direct training using a set of fixed weights over the objectives, does it perform better considering the training time and the optimality of the resulting Pareto front? 
To answer these questions, we prepared five continuous control tasks with several conflicting objectives in different simulated environments (Fig.~\ref{fig:overview}). 
We compared the results of our approach against the Radial Algorithm (RA) \cite{parisi2014policy} 
regarding the training time and the quality of the estimated Pareto front. 
For all of the tasks, we exploited the Proximal Policy Optimization (PPO) \cite{schulman2017proximal} algorithm as the policy learning method. 


\subsection{Environments}
The simulated environments are provided by Roboschool \cite{roboschool} and OpenAI gym \cite{1802.09464}.
The original environments return a single reward, which is the summation of rewards with respect to several objectives. Instead, we decompose the objectives to construct a reward vector. 
For each task, there is a minimum requirement that a policy needs to achieve in order to be considered as a valid policy (e.g., neither causing crashes nor being stuck at a robot joint configuration during the execution time). 
A detailed description of each task is presented in the followings.

\subsubsection{Reacher}
We start with a simple control task using a simulated arm with $2$ degrees-of-freedoms (DoFs). The goal is to move the tip of the arm as close as possible to a random target position. 
The environment returns a $3$-dimensional reward vector, which is related to the distance from the arm tip to the target position, an energy consumption cost and a stuck configuration penalty term. 
Among the $3$ rewards, the distance to the target and the energy consumption cost are conflicting.
A policy is considered valid if it receives an average discontented return of the stuck penalty less than $-0.05$.

\subsubsection{LunarLanderContinuous}
The goal of this task is to control the two engines of a rocket to land safely on the ground. The reward vector is a $4$-dimensional vector which includes (1) a shaping reward given when the rocket is upright and moving to the center of the landing area, (2) an energy consumption cost for the main engine, (3) an energy consumption cost for the side engine, and (4) a landing reward which can be positive or negative depending on the success or failure of the landing task.  
There is a weak conflict between the reward of shaping and the energy consumption cost for the engines. 
A policy is considered to be valid, i.e., not crashing, in case it receives an average discontented return of the landing reward greater than $20$.

\subsubsection{HalfCheetah, Ant and Humanoid}
To study the suitability of our MORL method to solve more
complex RL problems, we prepared a set of high-dimensional locomotion tasks using agents with different DoFs. 
The goal of the tasks is to control different agents to move forward. The agents are a half cheetah ($6$ DoFs), an ant ($8$ DoFs) and a humanoid robot ($17$ DoFs). All the agents use  5-dimensional reward vector consisting of (1) being alive, i.e., staying in an upright position, (2) gaining forward speed, (3) energy consumption penalty, (4) joint limit penalty, and (5) collision penalty. 
For every time step, the agent receives a fixed alive reward if it stays upright. The  forward speed reward is proportional to the  forward speed (negative when moves backward). The energy consumption penalty is  proportional to the amount of forces applied to all joints. The joint limit penalty punishes the agent  by a fixed negative value when a joint is positioned outside its limit. The collision cost punishes the agent for self collision. Among these rewards, the speed and the energy consumption rewards are conflicting, and cannot be optimized at the same time.
A policy is valid if it keeps the agent upright. This corresponds to an average discontented return of the alive reward greater than $80$ for the ant and the half cheetah, and $150$ for the humanoid.

\subsection{Results}
Here, we present the results of the proposed method to obtain the Pareto optimal policies and compare it with RA \cite{parisi2014policy} as the baseline. 
As described in Sec.~\ref{sec:relatedwork}, RA is a single-policy approach that estimates the Pareto front using multiple runs each of which solves a single-objective RL problem found by the scalarization function with different weights. The starting policy of RA is initialized randomly.
Our method work similar to RA but with the difference that, instead of randomly initializing the policy, the policy is initialized with the meta-policy.
The meta-policy is trained as explained in the previous section with $30$ random parameters of the scalarization functions. 

\subsubsection{Estimating the Pareto front}
We use the hypervolume indicator (\cite{van2013hypervolume}) to evaluate the quality of the Pareto fronts estimated by the two methods.
The hypervolume indicator calculates the volume encapsulated between a reference point and the Pareto front points. A larger hypervolume indicates a more optimal Pareto front.
The reference point is defined as the maximum value of each objective found by all policies.
The hypervolume of the Pareto front constructed by the two methods are listed in Table~\ref{table:hv}.
The results show that, for simple tasks, like LunarLander and Reacher, the performance of RA is equal or slightly better than our method. 
However for complex tasks, the proposed MORL method finds solutions resulting in larger hypervolumes, i.e., more optimal Pareto fronts.

To analyze the optimally of the Pareto front for each task, we only demonstrated three of the objectives in Fig.~\ref{fig:pf_plots}. We evaluated the dominance of each policy compared to all other policies and marked the non-dominated policies with black circles. The number of non-dominated policies found by the two methods is  given in Fig.~\ref{fig:non_dominated_p}. Considering both the hypervolume and the number of non-dominated policies measures confirms that our method outperforms the baseline in complex continuous control tasks. 

\subsubsection{Data-efficiency}
Here, we demonstrate the superiority of the proposed method to construct the Pareto front using less data. 
Fig.~\ref{fig:hv_indicator} illustrates the hypervolume measure calculated by our method during the fine-tuning phase. 
The red dashed lines denote the hypervolume found by the baseline method when all policies are converged.
To compare the data-efficiency of the methods, we observed how much data is used in each case.  
\newline
\noindent
\textbf{Baseline:} Each policy is trained with $1K$ iterations for Reacher and LunarLanderContinuous, $2K$ iterations for HalfCheetah and Ant, and $4K$ iterations for Humanoid.
In total 30 different policies are trained for each task. Therefore, as an example, for the Humanoid task $120K$ iterations are collected to obtain the hypervolume indicated by the red dashed line.  
\newline
\noindent
\textbf{Ours:}
The meta-policy is trained by the following number of meta-updates, i.e., the number of iterations for the adaptation and the meta-learning phases: $400$ for Reacher and LunarLanderContinuous, $200$ for HalfCheetah, $500$ for Ant and $690$ for Humanoid.
In our implementation, a meta-update requires five times more training data compared to a regular update. 
Therefore, as an example, with $(690\times5 + 70)\times 30 \sim 100K$ of iterations, our method reaches similar performance as achieved by $120K$ iterations of the baseline method. Furthermore, The proposed method keeps improving the results with more updates, i.e., almost $10\%$ improvement is achieved by $30K$ more iterations, while no more improvement can be observed by further training using the baseline method.

\subsubsection{The performance of individual policies}
We also studied the performance of each policy w.r.t. its corresponding single objective RL problem given by the scalarization function. 
For different scalarization functions, we compared our method and the baseline by enumerating the valid policies obtained by each method. 
Also, for each method, we counted the number of the policies that outperform the other method for the same scalar objective.
In this respect, 
the validity of a policy indicates whether the policy can accomplish a task and the accumulated scalar return indicates its performance for that specific task.
These measures are illustrated in Fig.~\ref{fig:valid_p} and Fig.~\ref{fig:higher_r}, respectively.
As it is shown, 
in both cases, the proposed method outperform the baseline, with greater margins for more complicated control tasks.

It is not surprising that the meta-learning MORL solution outperforms the baseline in this case. 
Assume that for the Humanoid task, we would like to learn a policy that minimizes the energy cost as the most important objective. In this case, training a randomly initialized policy will most likely result in a policy that always causing the agent to fall down. Since, this is the most likely local optimal behavior when the policy is initialized randomly. However, using meta-learning, the meta-policy is found such that the combination of different objectives can be achieved by few more training iterations. Therefore, in the above example, the policy can find a more optimal behavior that minimizes the energy consumption cost and still walks forward.  
In that sense, we can say that the meta-learning method results in a more efficient exploration strategy to acquire optimal behaviors. 

\begin{table}
\centering
\caption{Hypervolume indicator of the final Pareto front estimated by the two methods}
\begin{tabular}{ccc}
\hline
Environment & Our method            & RA             \\
\hline
Reacher     & 9.95          & \textbf{13.82}  \\
LunarLander & \textbf{453} & 422          \\
HalfCheetah & \textbf{$\mathbf{3.94\times10^6}$} & $3.64\times10^6$         \\
Ant         & \textbf{$\mathbf{10.4\times10^4}$} & $8.62\times10^4$          \\
Humanoid    & \textbf{$\mathbf{9.61\times10^6}$} & $8.78\times10^4$         \\
\hline
\end{tabular}
\label{table:hv}
\end{table}

\section{CONCLUSIONS}
\label{sec:conclusion}
In this work, we introduced a novel MORL approach based on meta-learning  to solve deep-MORL problems. We proposed to convert a MORL problem to a number of single-objective RL tasks using a parametric scalarization function. Then, using meta-learning, a meta-policy is obtained such that the performance on each task can be improved by few more training iterations.
The meta-policy is finally used as the optimal initial policy to train a set of Pareto optimal policies with different objectives.

We evaluated our method in several simulated continuous control tasks and demonstrated that it scales well to high-dimensional control problems. Furthermore, we demonstrated that it outperforms the only baseline method that can be applied to train deep policies to construct the Pareto front for reward vectors with several dimensions.   

\section{ACKNOWLEDGMENTS}
This work is supported by the European Union’s Horizon 2020 research and innovation program, the CENTAURO project (under grant agreement No. 644839), the  socSMCs project (H2020-FETPROACT-2014), and also by the Academy of Finland through the DEEPEN project. 

\addtolength{\textheight}{-12cm}  
\bibliographystyle{IEEEtran}
\balance

\begin{thebibliography}{10}
\providecommand{\url}[1]{#1}
\csname url@samestyle\endcsname
\providecommand{\newblock}{\relax}
\providecommand{\bibinfo}[2]{#2}
\providecommand{\BIBentrySTDinterwordspacing}{\spaceskip=0pt\relax}
\providecommand{\BIBentryALTinterwordstretchfactor}{4}
\providecommand{\BIBentryALTinterwordspacing}{\spaceskip=\fontdimen2\font plus
\BIBentryALTinterwordstretchfactor\fontdimen3\font minus
  \fontdimen4\font\relax}
\providecommand{\BIBforeignlanguage}[2]{{%
\expandafter\ifx\csname l@#1\endcsname\relax
\typeout{** WARNING: IEEEtran.bst: No hyphenation pattern has been}%
\typeout{** loaded for the language `#1'. Using the pattern for}%
\typeout{** the default language instead.}%
\else
\language=\csname l@#1\endcsname
\fi
#2}}
\providecommand{\BIBdecl}{\relax}
\BIBdecl

\bibitem{riedmiller2018learning}
M.~Riedmiller, R.~Hafner, T.~Lampe, M.~Neunert, J.~Degrave, T.~Van~de Wiele,
  V.~Mnih, N.~Heess, and J.~T. Springenberg, ``Learning by playing-solving
  sparse reward tasks from scratch,'' \emph{arXiv preprint arXiv:1802.10567},
  2018.

\bibitem{andrychowicz2017hindsight}
M.~Andrychowicz, F.~Wolski, A.~Ray, J.~Schneider, R.~Fong, P.~Welinder,
  B.~McGrew, J.~Tobin, O.~P. Abbeel, and W.~Zaremba, ``Hindsight experience
  replay,'' in \emph{Advances in Neural Information Processing Systems}, 2017,
  pp. 5048--5058.

\bibitem{ghadirzadeh2017deep}
A.~Ghadirzadeh, A.~Maki, D.~Kragic, and M.~Bj{\"o}rkman, ``Deep predictive
  policy training using reinforcement learning,'' in \emph{Intelligent Robots
  and Systems (IROS), 2017 IEEE/RSJ International Conference on}.\hskip 1em
  plus 0.5em minus 0.4em\relax IEEE, 2017, pp. 2351--2358.

\bibitem{schulman2017proximal}
J.~Schulman, F.~Wolski, P.~Dhariwal, A.~Radford, and O.~Klimov, ``Proximal
  policy optimization algorithms,'' \emph{arXiv preprint arXiv:1707.06347},
  2017.

\bibitem{duan2016benchmarking}
Y.~Duan, X.~Chen, R.~Houthooft, J.~Schulman, and P.~Abbeel, ``Benchmarking deep
  reinforcement learning for continuous control,'' in \emph{International
  Conference on Machine Learning}, 2016, pp. 1329--1338.

\bibitem{liu2015multiobjective}
C.~Liu, X.~Xu, and D.~Hu, ``Multiobjective reinforcement learning: A
  comprehensive overview,'' \emph{IEEE Transactions on Systems, Man, and
  Cybernetics: Systems}, vol.~45, no.~3, pp. 385--398, 2015.

\bibitem{vamplew2011empirical}
P.~Vamplew, R.~Dazeley, A.~Berry, R.~Issabekov, and E.~Dekker, ``Empirical
  evaluation methods for multiobjective reinforcement learning algorithms,''
  \emph{Machine learning}, vol.~84, no. 1-2, pp. 51--80, 2011.

\bibitem{roijers2013survey}
D.~M. Roijers, P.~Vamplew, S.~Whiteson, and R.~Dazeley, ``A survey of
  multi-objective sequential decision-making,'' \emph{Journal of Artificial
  Intelligence Research}, vol.~48, pp. 67--113, 2013.

\bibitem{finn2017model}
C.~Finn, P.~Abbeel, and S.~Levine, ``Model-agnostic meta-learning for fast
  adaptation of deep networks,'' \emph{arXiv preprint arXiv:1703.03400}, 2017.

\bibitem{parisi2014policy}
S.~Parisi, M.~Pirotta, N.~Smacchia, L.~Bascetta, and M.~Restelli, ``Policy
  gradient approaches for multi-objective sequential decision making,'' in
  \emph{Neural networks (ijcnn), 2014 international joint conference on}.\hskip
  1em plus 0.5em minus 0.4em\relax IEEE, 2014, pp. 2323--2330.

\bibitem{miettinen2002scalarizing}
K.~Miettinen and M.~M. M{\"a}kel{\"a}, ``On scalarizing functions in
  multiobjective optimization,'' \emph{OR spectrum}, vol.~24, no.~2, pp.
  193--213, 2002.

\bibitem{miettinen2001some}
K.~Miettinen, ``Some methods for nonlinear multi-objective optimization,'' in
  \emph{International Conference on Evolutionary Multi-Criterion
  Optimization}.\hskip 1em plus 0.5em minus 0.4em\relax Springer, 2001, pp.
  1--20.

\bibitem{coello2007evolutionary}
C.~A.~C. Coello, G.~B. Lamont, D.~A. Van~Veldhuizen \emph{et~al.},
  \emph{Evolutionary algorithms for solving multi-objective problems}.\hskip
  1em plus 0.5em minus 0.4em\relax Springer, 2007, vol.~5.

\bibitem{konak2006multi}
A.~Konak, D.~W. Coit, and A.~E. Smith, ``Multi-objective optimization using
  genetic algorithms: A tutorial,'' \emph{Reliability Engineering \& System
  Safety}, vol.~91, no.~9, pp. 992--1007, 2006.

\bibitem{zitzler2001spea2}
E.~Zitzler, M.~Laumanns, and L.~Thiele, ``Spea2: Improving the strength pareto
  evolutionary algorithm,'' \emph{TIK-report}, vol. 103, 2001.

\bibitem{deb2000fast}
K.~Deb, S.~Agrawal, A.~Pratap, and T.~Meyarivan, ``A fast elitist non-dominated
  sorting genetic algorithm for multi-objective optimization: Nsga-ii,'' in
  \emph{International Conference on Parallel Problem Solving From
  Nature}.\hskip 1em plus 0.5em minus 0.4em\relax Springer, 2000, pp. 849--858.

\bibitem{brys2017multi}
T.~Brys, A.~Harutyunyan, P.~Vrancx, A.~Now{\'e}, and M.~E. Taylor,
  ``Multi-objectivization and ensembles of shapings in reinforcement
  learning,'' \emph{Neurocomputing}, vol. 263, pp. 48--59, 2017.

\bibitem{lizotte2010efficient}
D.~J. Lizotte, M.~H. Bowling, and S.~A. Murphy, ``Efficient reinforcement
  learning with multiple reward functions for randomized controlled trial
  analysis,'' in \emph{Proceedings of the 27th International Conference on
  Machine Learning (ICML-10)}.\hskip 1em plus 0.5em minus 0.4em\relax Citeseer,
  2010, pp. 695--702.

\bibitem{das1997closer}
I.~Das and J.~E. Dennis, ``A closer look at drawbacks of minimizing weighted
  sums of objectives for pareto set generation in multicriteria optimization
  problems,'' \emph{Structural optimization}, vol.~14, no.~1, pp. 63--69, 1997.

\bibitem{van2013scalarized}
K.~Van~Moffaert, M.~M. Drugan, and A.~Now\'e, ``Scalarized multi-objective
  reinforcement learning: Novel design techniques.'' in \emph{ADPRL}, 2013, pp.
  191--199.

\bibitem{van2013hypervolume}
K.~Van~Moffaert, M.~M. Drugan, and A.~Now{\'e}, ``Hypervolume-based
  multi-objective reinforcement learning,'' in \emph{International Conference
  on Evolutionary Multi-Criterion Optimization}.\hskip 1em plus 0.5em minus
  0.4em\relax Springer, 2013, pp. 352--366.

\bibitem{natarajan2005dynamic}
S.~Natarajan and P.~Tadepalli, ``Dynamic preferences in multi-criteria
  reinforcement learning,'' in \emph{Proceedings of the 22nd international
  conference on Machine learning}.\hskip 1em plus 0.5em minus 0.4em\relax ACM,
  2005, pp. 601--608.

\bibitem{handa2009eda}
H.~Handa, ``Eda-rl: estimation of distribution algorithms for reinforcement
  learning problems,'' in \emph{Proceedings of the 11th Annual conference on
  Genetic and evolutionary computation}.\hskip 1em plus 0.5em minus 0.4em\relax
  ACM, 2009, pp. 405--412.

\bibitem{handa2009solving}
------, ``Solving multi-objective reinforcement learning problems by
  eda-rl-acquisition of various strategies,'' in \emph{2009 ninth international
  conference on intelligent systems design and applications}.\hskip 1em plus
  0.5em minus 0.4em\relax IEEE, 2009, pp. 426--431.

\bibitem{pirotta2015multi}
M.~Pirotta, S.~Parisi, and M.~Restelli, ``Multi-objective reinforcement
  learning with continuous pareto frontier approximation,'' in \emph{29th AAAI
  Conference on Artificial Intelligence, AAAI 2015 and the 27th Innovative
  Applications of Artificial Intelligence Conference, IAAI 2015}.\hskip 1em
  plus 0.5em minus 0.4em\relax AAAI Press, 2015, pp. 2928--2934.

\bibitem{parisi2017manifold}
S.~Parisi, M.~Pirotta, and J.~Peters, ``Manifold-based multi-objective policy
  search with sample reuse,'' \emph{Neurocomputing}, vol. 263, pp. 3--14, 2017.

\bibitem{schulman2015trust}
J.~Schulman, S.~Levine, P.~Abbeel, M.~Jordan, and P.~Moritz, ``Trust region
  policy optimization,'' in \emph{International Conference on Machine
  Learning}, 2015, pp. 1889--1897.

\bibitem{roboschool}
``Roboschool: open-source software for robot simulation,''
  \url{https://blog.openai.com/roboschool/}.

\bibitem{1802.09464}
M.~Plappert, M.~Andrychowicz, A.~Ray, B.~McGrew, B.~Baker, G.~Powell,
  J.~Schneider, J.~Tobin, M.~Chociej, P.~Welinder, V.~Kumar, and W.~Zaremba,
  ``Multi-goal reinforcement learning: Challenging robotics environments and
  request for research,'' 2018.

\end{thebibliography}

\end{document}